\documentclass[journal]{IEEEtran}

\usepackage{algorithm}
\usepackage[noend]{algpseudocode}
\usepackage{amsmath,amsfonts}
\usepackage{balance}
\usepackage{booktabs}
\usepackage{graphicx}
\usepackage[hidelinks]{hyperref}
\usepackage{stfloats}
\usepackage{subcaption}
\usepackage{textcomp}
\usepackage{tikz}
\usepackage{url}
\usepackage{xcolor}

\usepackage[backend=biber,hyperref=true,backref=true,citestyle=numeric-comp,sorting=none]{biblatex}
\def\BibTeX{{\rm B\kern-.05em{\sc i\kern-.025em b}\kern-.08em
    T\kern-.1667em\lower.7ex\hbox{E}\kern-.125emX}}

\begin{document}

    \title{Single-pass Possibilistic Clustering with Damped Window Footprints}

\author{
    \IEEEauthorblockN{
        Jeffrey Dale,
        James Keller,~\IEEEmembership{Life Fellow, IEEE},
        Aquila Galusha
    } \\
    \IEEEauthorblockA{
        Electrical Engineering and Computer Science Department \\
        University of Missouri, Columbia, MO 65211
    }
    \thanks{
        This research was conducted as part of the first author's Ph.D. dissertation from 2018--2022, entitled
        \textit{Building Environmentally-Aware Classifiers on Streaming Data}. \\
        \textit{Corresponding author: Jeffrey Dale (jjdale@mail.missouri.edu)}
    }
}

\maketitle

\begin{abstract}
    Streaming clustering is a domain that has become extremely relevant in the age of big data, such as in network traffic analysis or in processing
        continuously-running sensor data.
    Furthermore, possibilistic models offer unique benefits over approaches from the literature, especially with the introduction of a ``fuzzifier''
        parameter that controls how quickly typicality degrades as one gets further from cluster centers.
    We propose a single-pass possibilistic clustering (SPC) algorithm that is effective and easy to apply to new datasets.
    Key contributions of SPC include the ability to model non-spherical clusters, closed-form footprint updates over arbitrarily sized damped windows,
        and the employment of covariance union from the multiple hypothesis tracking literature to merge two cluster mean and covariance estimates.
    SPC is validated against five other streaming clustering algorithm on the basis of cluster purity and normalized mutual information.
\end{abstract}

    \section{Introduction}

\IEEEPARstart{T}{he} formulation of the streaming clustering problem varies depending on the source.
In general, there is agreement that points from a data stream cannot be retained, they must be processed and then discarded.
While this requirement may seem arbitrary in a time where buying more hardware is cheaper than improving methodology, certain data sources
    (\textit{e.g.}, network traffic) simply provide so much information at such a high rate that the time and memory resources needed to iterate over
    the data are astronomical.
Thus, there is high demand for streaming data analysis (SDA) algorithms, that is, algorithms that only make a single pass over the data.

We have observed a lack of possibilistic approaches to streaming clustering in the literature, which we believe (and will show) to have numerous
    desirable properties to offer.
In this paper, we propose a single-pass possibilistic clustering (SPC) algorithm for the task of streaming clustering.
As the name suggests, SPC adopts a possibilistic model, but with modifications that enable it to detect arbitrarily-shaped clusters.

SPC maintains a set of $n$ structures that are intelligently placed in feature space so as to best describe the data stream.
When a new point arrives, it is given its own structure and then, to keep the number of structures constant, the two most compatible structures are
    merged with respect to a Mahalanobis distance-based typicality measure.
The priority given to recent points can be adjusted with decay factor parameters $\gamma$ and $\beta$.
To obtain a clustering of the stream, DBSCAN~\cite{ester1996density} is employed using a distance measure derived from SPC's typicality
    measure.

We evaluate the proposed SPC on stationary and non-stationary datasets of varying dimensionality and show that SPC achieves very high performance
    metrics in all cases, consistently either outperforming or staying competitive with related algorithms.
As a qualitative evaluation, we also show that the decision region induced by SPC structures on two dimensional data is accurate to what human
    intuition would suggest.

The rest of the paper is organized as follows.
Section~\ref{sec:related_work} identifies related algorithms to SPC in the literature and details their similarities to and differences from SPC.
Next, the SPC algorithm is described in detail in Section~\ref{sec:methods}, with the full algorithm presented in Algorithm~\ref{alg:spc}.
SPC is then evaluated against five state of the art streaming clustering algorithms on several datasets in Section~\ref{sec:experiments} and
    conclusions about SPC are drawn in Section~\ref{sec:conclusion}.
    \section{Related Work}
\label{sec:related_work}

Few approaches in the past have sought to obtain a meaningful possibilistic clustering from data streams, though notable methods include Hu et al.
    with an evolutionary approach~\cite{hu2018neuro} and Wu et al. with a neural gas-based approach~\cite{wu2021streamsong}.
Additionally, Škrjanc et al. study the use of various norm-inducing matrices for evolving possibilistic clustering in data
    streams~\cite{vskrjanc2019inner}, including the inverse covariance matrix (\textit{i.e.}, Mahalanobis distance) which we employ in this work.

It is common for algorithms that operate on streaming data to employ some kind of windowing model to allow the algorithm to focus on newer and
    potentially more relevant data points.
The most used windowing models in the literature are sliding windows, damped windows, and pyramidal windows.
The general idea of sliding windows is to only consider a fixed number of recent data points, whereas all points in the stream have an impact in
    damped and pyramidal window approaches.
Pyramidal windows attempt to summarize the entire stream with a finite amount of information by maintaining coarser detail based on how far back
    points are in the stream.
As damped windowing is employed in this work, we describe it in greater detail in this section.

The idea of damping has been studied for centuries in the form of damped motion in physics.
The first use of damped windows in streaming clustering seems to be with DenStream~\cite{cao2006density} as a mechanism to determine when
    micro-clusters should be discarded.
Since then, as Zubaroğlu et al.~\cite{zubarouglu2021data} point out, a number of algorithms have used damped
    windowing~\cite{hahsler2016clustering,de2017evolutionary,amini2016mudi,hyde2017fully,bechini2020tsf} in some capacity.

Many streaming clustering algorithms (\textit{e.g.}, 
    \cite{cao2006density,hahsler2016clustering,bechini2020tsf,de2017evolutionary,amini2016mudi,hyde2017fully}), including SPC, use the idea of damped
    windows to assign a weight to previously seen points in the stream, usually as a means to determine when to discard information about these
    points.
Less commonly, damped window mean estimates are used, and applying a damped window model to covariance estimation is novel to
    streaming clustering.

From the crisp streaming clustering literature, DBSTREAM~\cite{hahsler2016clustering} has the most overlap with SPC, primarily due to both algorithms'
    dependence on DBSCAN~\cite{ester1996density} and damped windows.
Key differences are that DBSTREAM bases its streaming updates on ideas from DBSCAN, whereas SPC uses DBSCAN directly for offline clustering, albeit
    with a specialized distance function.
As for damped windows, DBSTREAM is one of many algorithms to use them in determining cluster weights.

Bechini et al. have proposed TSF-DBSCAN~\cite{bechini2020tsf} which also shares some ideas with SPC.
Both algorithms use DBSCAN, however TSF-DBSCAN uses a fuzzy extension of DBSCAN as an offline clustering procedure and SPC uses DBSCAN with a
    typicality-aware distance function.
Again, both algorithms, like many others, use damped windows for measuring the weight of a structure.
The most notable, and fundamental, difference is that TSF-DBSCAN is built on a fuzzy framework and SPC is built on a possibilistic framework.
    \section{Methodology}
\label{sec:methods}

We have divided the methodology of SPC into three sections.
In Section~\ref{ssec:methods:typicality}, we motivate the need for a possibilistic model as opposed to probabilistic model for streaming clustering,
    and then describe the proposed possibilistic model.
In Section~\ref{ssec:methods:footprints}, we explain how the structures used in SPC can be represented with fixed-size footprints and how to perform
    relevant operations using these footprints.
Then, in Section~\ref{ssec:methods:cu}, we outline the method of covariance union~\cite{julier2004method} from the domain of multiple hypothesis
    tracking and how it is valuable in streaming clustering.
Finally, Section~\ref{ssec:methods:description} contains an intuitive description of the SPC algorithm with the full pseudocode provided in
    Algorithm~\ref{alg:spc}.

\subsection{Typicality with Mahalanobis Distance}
\label{ssec:methods:typicality}

When using a mean and covariance matrix to model a structure, the natural method of determining the degree to which an arbitrary point belongs to
    the structure is to treat it as a Gaussian distribution and use the PDF to compute the probability that the point may have come from it.
However, a possibilistic approach introduces a valuable ``fuzzifier'' parameter that controls how quickly typicality degrades when getting further
    from a structure.
To compute the typicality of a point $x$ within a structure $s$, the general equation, originally presented by Krishnapuram and
    Keller~\cite{krishnapuram1993possibilistic}, is given by
\begin{align} \label{eq:typicality_pcm}
    u_m(s, x) = \frac{1}{1 + \left( \frac{d(s, x)^2}{\eta} \right) ^ \frac{1}{m-1}}
\end{align}
where $m > 1$ is a fuzzifier parameter, $d(s, x)$ is a distance measure between the structure $s$ and a point $x$, and $\eta$ is a ``suitable
    positive number'' that acts as a scale parameter governing the (squared) distance at which typicality reaches $1/2$.

In the possibilistic $C$-means algorithm (PCM)~\cite{krishnapuram1993possibilistic} where Equation~\ref{eq:typicality_pcm} was presented, the distance
    was defined to be Euclidean, thus constraining PCM to only find hyperspherical clusters.
Replacing Euclidean distance with Mahalanobis distance leads to a class of typicality measures on $m$ given by
\begin{align} \label{eq:mahal_typicality}
    u_m(x, \mu, \Sigma) = \frac{1}{1 + \left[ (x - \mu)^T \Sigma^{-1} (x - \mu) \right]^\frac{1}{m-1}}.
\end{align}
where $\mu$ is a structure mean, $\Sigma$ is a measure of structure covariance, and $m$ is the fuzzifier parameter as before.
When $\Sigma = \eta \mathbb{I}$, Equation~\ref{eq:mahal_typicality} reduces to Equation~\ref{eq:typicality_pcm}.
Generalizing $\eta$ to be a covariance matrix allows for the detection of hyperellipsoidal clusters as in the possibilistic Gustafson-Kessel (PGK)
    algorithm~\cite{krishnapuram1993possibilistic}.

It should be noted that Equation~\ref{eq:mahal_typicality} no longer satisfies the necessary conditions for convergence of the optimization procedure
presented by Krishnapuram and Keller~\cite{krishnapuram1993possibilistic}, but, in this work, we are using an entirely different method for
finding structure in the data stream.

\begin{figure}[t!]
     
    \centering

    \begin{subfigure}{0.47\textwidth}
        \centering
        \includegraphics[width=0.9\textwidth]{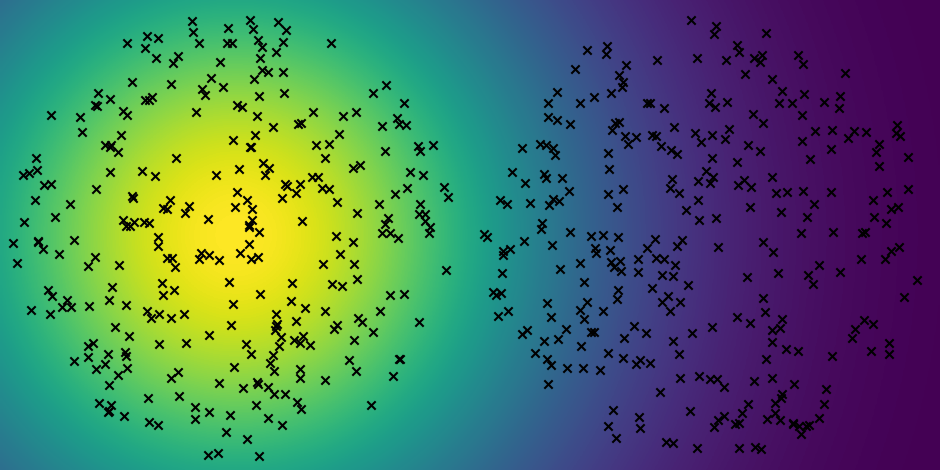}
        \caption{
            A Gaussian model assigns high probability in some points in the right circle.
        }
    \end{subfigure}
    \hspace{1em}
    \begin{subfigure}{0.47\textwidth}
        \centering
        \includegraphics[width=0.9\textwidth]{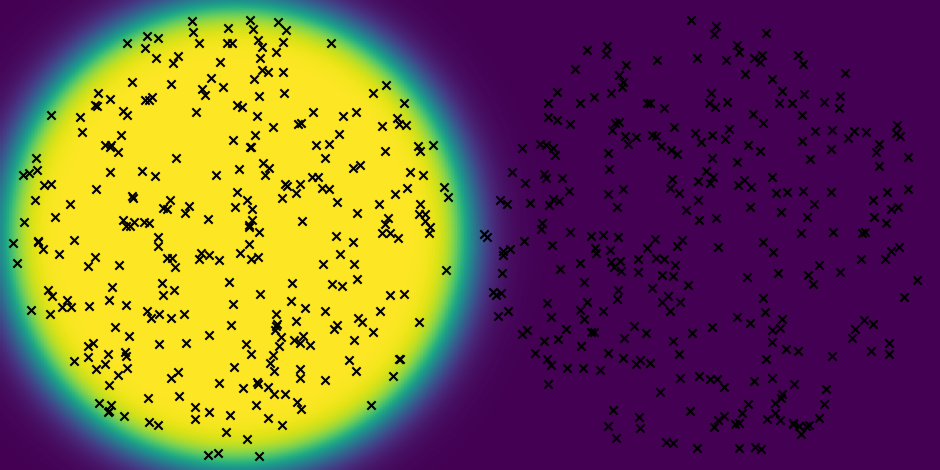}
        \caption{
            The fuzzifier parameter $m$ in the possibilistic model allows fine control over typicality falloff, effectively separating the two
            circles.
            Here, we set $m=1.1$.
        }
    \end{subfigure}

    \caption{
        Motivation for using a possibilistic model over a probabilistic model in SPC demonstrated on a synthetic dataset of two near-overlapping
        circles.
    }
    
   \label{fig:mahal_motivation}

\end{figure}

The introduction of the fuzzifier parameter in possibilistic models is especially useful when clusters in the data are very close together, but not
    overlapping.
A synthetic dataset that illustrates this scenario is shown in Figure~\ref{fig:mahal_motivation}.
Choosing the fuzzifier parameter $m$ to be small allows the possibilistic model to tightly cover the left circle without allowing points in the right
    circle to have high typicality in it.
A Gaussian model, for example, cannot cover the left circle without assigning high probability to points in the right circle as well.

A key measure that is utilized in SPC is the distance between two structures $s_1$ and $s_2$.
This distance must take into account the mean and covariance of both structures, and we would be discarding important information to simply compute
    the Euclidean distance between their means.
We instead utilize the typicality measure of Equation~\ref{eq:mahal_typicality} and transform it into a distance measure according to
\begin{equation} \label{eq:structure_distance}
    D(s_1, s_2) = 1 - u_m(\mu_2, \mu_1, \Sigma_1) u_m(\mu_1, \mu_2, \Sigma_2).
\end{equation}
The product of the typicality of $\mu_2$ in $s_1$ and $\mu_1$ in $s_2$ is used to make the measure symmetric, and the subtraction from one turns it
    from a measure of similarity to dissimilarity.
The utilization of typicalities in the distance measure of Equation~\ref{eq:structure_distance} is crucial to the performance of SPC, as it
    determines when structures are merged, deleted, and which structure are clustered together with DBSCAN.
Thus, even though outputs of SPC are crisp, the algorithm relies heavily on concepts from possibility theory, especially those proposed in PCM.

It is also valuable to transform typicalities from Equation~\ref{eq:mahal_typicality} into a logarithmic scale for the purpose of specifying
    parameter values in more intuitive ranges.
To this end, we define the negative log typicality (NLT) as
\begin{align} \label{eq:nlt}
    NLT(x, \mu, \Sigma) = -\log u_m(x, \mu, \Sigma),
\end{align}
where, in this work, $\log$ refers to the natural logarithm.
Like the negative log likelihood (NLL) in statistics, NLT takes on the range $[0, \infty)$ with higher values reflecting lower typicalities.
For our experiments, we use an NLT threshold of 3 (see Table~\ref{tab:parameters}), corresponding to a typicality of around 0.05.

\subsection{Structure Footprints}
\label{ssec:methods:footprints}

Each structure at time $T$ is represented by its mean $\mu^{(T)}$ and a symmetric positive-definite (SPD) measure of spread $\Sigma^{(T)}$ that
    behaves like a covariance matrix.
These measures are computed according to a damped window approach with decay factor $\gamma \ge 0$ that assigns exponentially decaying weight to older
    observations.

In general, damped window approaches apply an exponential coefficient such as $2^{-\gamma (T-t)}$ to the ``importance'' at the current time $T$ of a
    point that arrived at time $t$.
The parameter $\gamma$ here serves as a damping factor that models how long it takes for points to be discounted.
While, in theory, every point in an infinitely long stream would have some contribution in this model, this contribution very quickly becomes zero
    when working with fixed-precision numerical representations.
For this reason, the damping factor $\gamma$ is usually picked to be very small, \textit{e.g.}, $10^{-3}$.

In addition to tracking a mean and covariance for each structure, the structure footprint includes a measure of the average weight $w^{(T)}$ in this
    structure which determines how much typicality recent points in the stream have had in this structure, much like the idea of microcluster weights
    in DenStream and DBSTREAM.
Since it is sometimes useful to apply a larger window to structure weights than for mean and covariance updates, the structure weight uses a separate
    decay factor $\beta \ge 0$ the behaves exactly as $\gamma$.

The footprint for a structure, consisting of mean $\mu^{(T)}$, spread $\Sigma^{(T)}$, and weight $w^{(T)}$ at time $T$, is computed according to
\begin{align}
    \label{eq:static:mu}
    \mu^{(T)}    &= \frac{1}{\Gamma^{(T)}} \sum_{t=1}^T e^{-\gamma (T - t)} x^{(t)} \\
    \label{eq:static:sigma}
    \Sigma^{(T)} &= \frac{1}{\Gamma^{(T)}} \sum_{t=1}^T e^{-\gamma (T - t)} (x^{(t)} - \mu^{(t)})(x^{(t)} - \mu^{(t)})^T \\
    \label{eq:static:w}
    w^{(T)}      &= \frac{1}{B^{(T)}}      \sum_{t=1}^T e^{-\beta  (T - t)} u_m(x^{(t)}, \mu^{(T)}, \Sigma^{(T)}) \\
    \label{eq:static:gamma}
    \Gamma^{(T)} &= \sum_{t=1}^T e^{-\gamma (T-t)} 
                  = \begin{cases}
                        \frac{e^\gamma - e^{\gamma(1-T)}}{e^\gamma - 1}, & \gamma > 0 \\
                        T, & \gamma = 0
                    \end{cases} \\
    \label{eq:static:beta}
    B^{(T)}      &= \sum_{t=1}^T e^{-\beta (T-t)}
                 = \begin{cases}
                        \frac{e^\beta - e^{\beta(1-T)}}{e^\beta - 1}, & \beta > 0 \\
                        T, & \beta = 0
                    \end{cases}
\end{align}
where $x^{(t)}$ represents the $t$-th observation in the stream, $u_m$ is as in Equation~\ref{eq:mahal_typicality}, and $\Gamma^{(T)}$ and $B^{(T)}$
    are normalizing coefficients, differing only in their decay factor.
We notice that when $\gamma = 0$, $\mu^{(T)}$ reduces to the mean of the first $T$ elements of the stream and $\Gamma^{(T)} = T$.
Thus, the damped window mean is a generalization of the usual arithmetic mean.
Similarly, when $\gamma = 0$, the damped window covariance reduces to (biased) sample covariance.

We can merge two structures with footprints $(\mu_1^{(T_1)}, \Sigma_1^{(T_1)}, w^{(T_1)})$ and $(\mu_2^{(T_2)}, \Sigma_2^{(T_2)}, w^{(T_2)})$ quite
    easily with the following equations:
\begin{align}
    \label{eq:merge:mu}
    \mu^{(T)}    &= \frac{1}{\Gamma^{(T_1 + T_2)}} \left( e^{-\gamma T_2} \Gamma^{(T_1)} \mu_1^{(T_1)} + \Gamma^{(T_2)} \mu_2^{(T_2)} \right) \\
    \label{eq:merge:sigma}
    \Sigma^{(T)} &= \frac{1}{\Gamma^{(T_1 + T_2)}} \left( e^{-\gamma T_2} \Gamma^{(T_1)} \Sigma_1^{(T_1)} + \Gamma^{(T_2)} \Sigma_2^{(T_2)} \right) \\
    \label{eq:merge:w}
    w^{(T)}      &= \frac{1}{B^{(T_1 + T_2)}} \left( e^{-\beta T_2} B^{(T_1)} w_1^{(T_1)} + B^{(T_2)} w_2^{(T_2)} \right)
\end{align}
where $T = T_1 + T_2$.
The multiplication of $\mu^{(T_1)}$ by $\Gamma^{(T_1)}$ in Equation~\ref{eq:merge:mu} inverts the normalization done in the computation of
    $\mu^{(T_1)}$ and the multiplication by $e^{-\gamma T_2}$ effectively shifts the weight of each $x^{(t)}$ involved in computing $\mu^{(T_1)}$
    back $T_2$ units of time.
By the same logic, inverting the normalization done in computing $\mu_2^{(T_2)}$ is done by multiplication with $\Gamma^{(T_2)}$.
Then, the two terms can be added together and re-normalized by $\Gamma^{(T)}$, which one will find is equivalent to
    Equation~\ref{eq:static:mu}.
The intuition behind the incremental updates for $\Sigma^{(T)}$ and $w^{(T)}$ is identical.
Note that, while $\Gamma^{(T)}$ and $B^{(T)}$ can also be updated incrementally, they can more easily be computed in closed form
    as shown in Equations~\ref{eq:static:gamma} and~\ref{eq:static:beta}.

We also make use of the ability to update structure weight incrementally given a new point $x$ according to
\begin{align} \label{eq:incremental:w}
    w^{(T+1)} &= \frac{1}{B^{(T+1)}} \left( e^{-\beta} B^{(T)} w^{(T)} + u_m(x, \mu^{(T)}, \Sigma^{(T)}) \right).
\end{align}
We observe Equation~\ref{eq:incremental:w} to again follow the same general form as the other damped window updates.

All methods of computing structure footprints $(\mu^{(T)}, \Sigma^{(T)})$ are equivalent in theory, however for numerical stability, one may be
    concerned that repeated multiplication by $\Gamma^{(T-1)}$ and division by $\Gamma^{(T)}$ will lead to floating point errors that propagates
    indefinitely over time.
If one was so inclined, they could instead maintain un-normalized estimates of $\mu^{(T)}$ and $\Sigma^{(T)}$ and only normalize by $\Gamma^{(T)}$
    when needed.
This is much akin to the summary statistics of the sum of data points and the sum of squared data points that are commonly used in incremental
    estimates of mean and covariance, but generalized to conform to the damped window model.

\subsection{Covariance Union}
\label{ssec:methods:cu}

\begin{figure}[t!]
     
    \centering

    \begin{subfigure}{0.47\textwidth}
        \centering
        \includegraphics[width=0.75\textwidth]{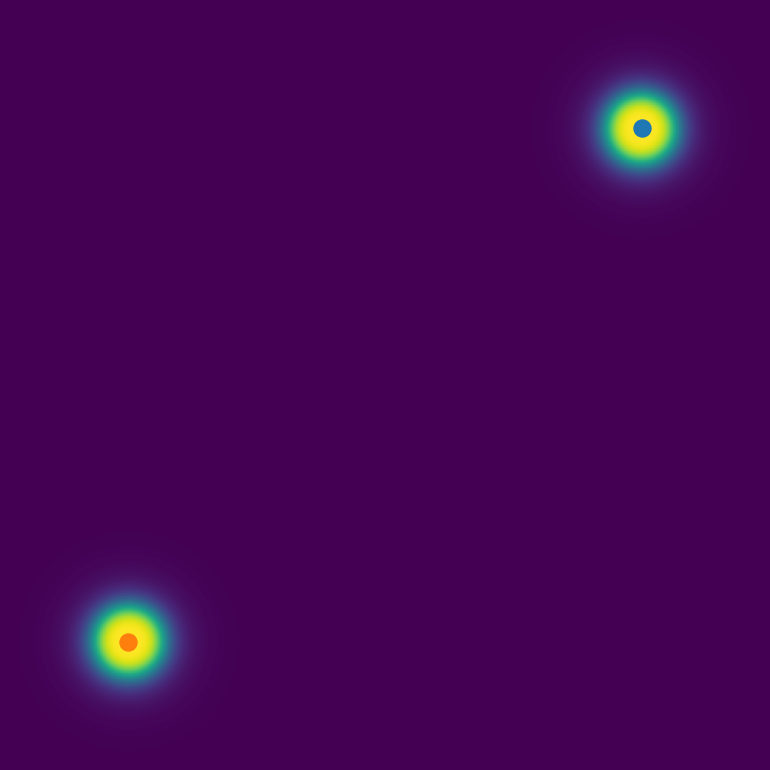}
        \caption{
                Two example structures with relatively small covariances that we want to merge.
        }
    \end{subfigure}
    
    \vspace{1em}

    \begin{subfigure}{0.47\textwidth}
        \centering
        \includegraphics[width=0.75\textwidth]{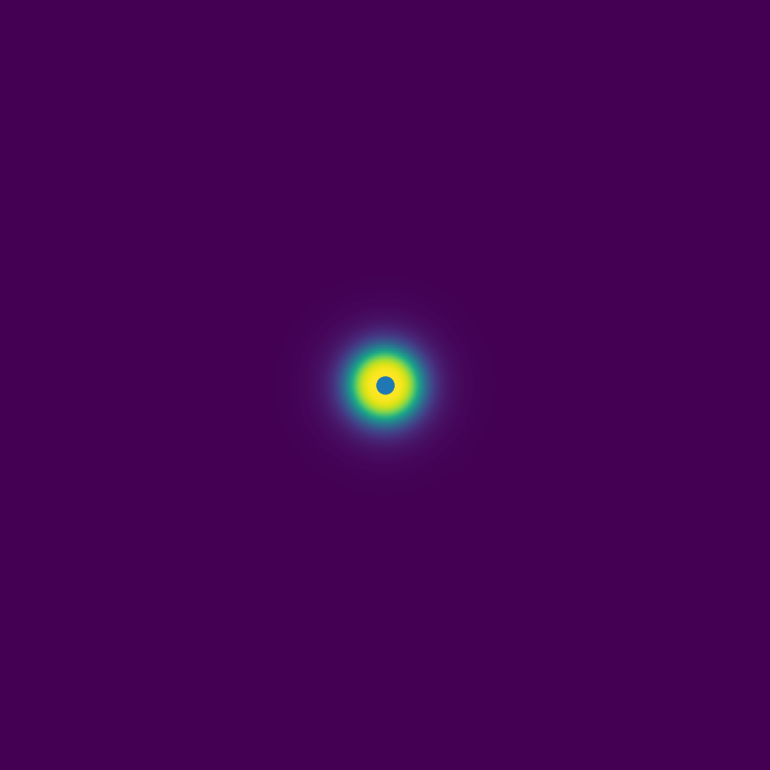}
        \caption{
                Result of merging two structures using Equation~\ref{eq:merge:sigma}.
        }
    \end{subfigure}

    \vspace{1em}

    \begin{subfigure}{0.47\textwidth}
        \centering
        \includegraphics[width=0.75\textwidth]{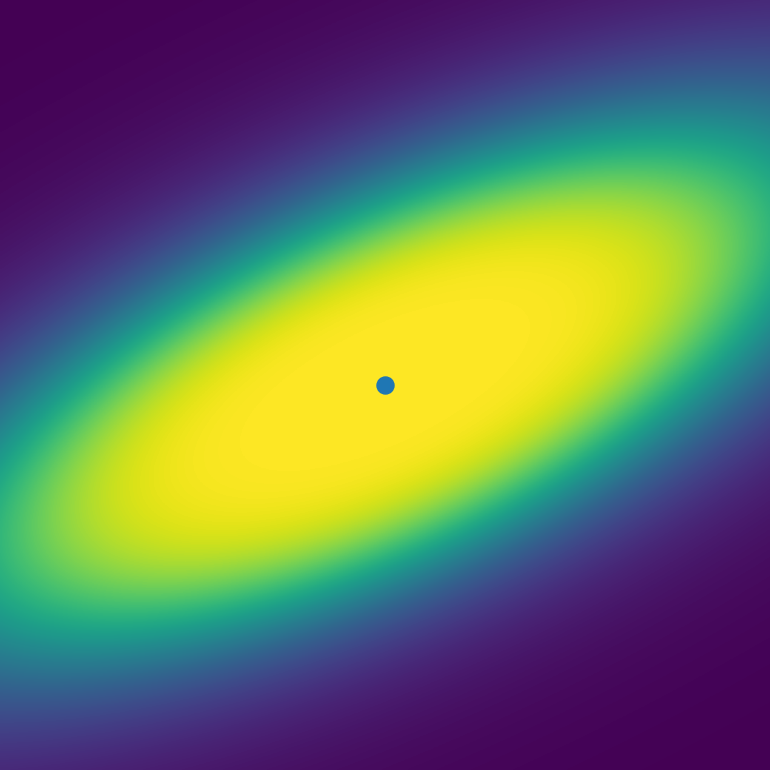}
        \caption{
                Result of merging two structures using covariance union.
        }
    \end{subfigure}
   
    \caption{
        Illustration of why covariance union is needed to combined covariance matrices of two structures with unequal means.
    }
    
   \label{fig:cu_motivation}

\end{figure}

The method of fusing the covariance of two structures presented in Equation~\ref{eq:merge:sigma} is only valid when we are considering two structures
    with the same mean.
When the means of two structures $s_1$ and $s_2$ are different, we want a new covariance matrix that is large enough to encompass the region of
    feature space influenced by both constituent structures, which, in the event that the means of $s_1$ and $s_2$ are far apart, can be much larger
    than either of the individual covariances.
Figure~\ref{fig:cu_motivation} demonstrates why this is the case.

The problem of fusing two or more unique mean and covariance pairs has been studied extensively in the filtering literature, specifically in the
    domain of multiple hypothesis tracking.
In this work, we employ covariance union (CU)~\cite{julier2004method} for obtaining the fused covariance of two merged structures.
It should be noted that CU was developed in the context of ensuring fused covariances are conservative estimators of the true state of the object(s)
    being tracked, but here, as we are not tracking anything per se, the method is used due simply to its observed effectiveness at producing high
    quality fused covariance estimates in our experiments.

Suppose again that we have two structures $s_1$ and $s_2$ with footprints $(\mu_1^{(T_1)}, \Sigma_1^{(T_1)}, w^{(T_1)})$ and
    $(\mu_2^{(T_2)}, \Sigma_2^{(T_2)}, w^{(T_2)})$ that we wish to fuse.
The first step of CU is to determine a candidate mean $\mu^{(T)}$, where $T = T_1 + T_2$, of the fused state of the system, which we have established can be done with
    Equation~\ref{eq:merge:mu}.
We then define two matrices $U_1$ and $U_2$ to be covariances $\Sigma_1$ and $\Sigma_2$ of $s_1$ and $s_2$ padded by the outer product of the
    difference between the structure mean and the candidate mean according to
\begin{align}
    U_1 &= \Sigma_1 + \left( \mu^{(T)} - \mu^{(T_1)} \right) \left( \mu^{(T)} - \mu^{(T_1)} \right)^T \\
    U_2 &= \Sigma_2 + \left( \mu^{(T)} - \mu^{(T_2)} \right) \left( \mu^{(T)} - \mu^{(T_2)} \right)^T.
\end{align}

Then, let $U_2 = LL^T$ be the Cholesky decomposition of $U_2$, where $L$ is lower triangular, and compute the eigendecomposition
    $L^{-T}U_1L^{-1} = Q \Lambda Q^T$, where the columns of $Q$ are orthonormal eigenvectors and $\Lambda$ is a diagonal matrix of eigenvalues of
    $L^{-T}U_1L^{-1}$.
The fused covariance estimate $\Sigma^{(T)}$ is thus given by
\begin{align} \label{eq:cu}
    \Sigma^{(T)} = LQ \max\left\{ \Lambda, \mathbb{I} \right\} Q^T L^T
\end{align}
where $\max$ is taken elementwise between its arguments.
Proof that the covariance matrix resulting from Equation~\ref{eq:cu} is conservative is provided in the original manuscript~\cite{julier2004method},
    though, to reiterate, our inputs to Equation~\ref{eq:cu} do not, in general, satisfy the assumptions required for that result to hold.

\begin{algorithm}[ht!]
    \caption{Single-pass Possibilistic Clustering}
    \label{alg:spc}
    
        \begin{algorithmic}

            \State \textit{// Let $\mu_i$ be the mean of the $i$-th structure}
            \State \textit{// Let $\Sigma_i$ be the covariance of the $i$-th structure}
            \State \textit{// Let $w_i$ be the weight of the $i$-th structure}
            \State \textit{// Let $T_i$ be the number of timesteps elapsed since the $i$-th}
            \State \textit{// \hspace{1em} structure was created}
            \State \textit{// Let $S = \{ (\mu_i, \Sigma_i, w_i, T_i) \}$ be the set of $n$ structures}
            \State \textit{// \hspace{1em} currently tracked by SPC, initially empty}
            \State \textit{// Let $N$ be the maximum number of structures to track}
            \State
    
            \Procedure{UpdateSPC}{$x$, $m$, $\gamma$, $\beta$}

                \State
                \State \textit{// Create a new structure to accommodate this point.}
                \State $S \gets S \cup \{ ( x, \mathbb{I}, 1, 1 ) \}$
                \State

                \State \textit{// If there are too many structures, prune and merge.}
                \If{$\left\lvert S \right\rvert > N$}
                    \State
                    \For{$i = 1, 2, \dots, |S|$}
                        \State Update $w_i$ using Equation~\ref{eq:incremental:w}.
                        \If{$w_i < w_{min}$}
                            \State Let $\hat{j} = \underset{j = 1, \dots, |S|, \ j \ne i}{argmin} \ NLT(\mu_i, \mu_j \Sigma_j)$.
                            \If{$NLT(\mu_i, \mu_{\hat{j}} \Sigma_{\hat{j}}) < NLT_{max}$}
                                \State Call \texttt{Merge}($\hat{j}$, $i$).
                            \EndIf
                            \State Delete $s_i$ from $S$.
                        \EndIf
                    \EndFor

                    \State
                    \State \textit{// Find two closest structures and merge them.}
                    \State \textit{// $D(i, j)$ is computed as in Equation~\ref{eq:structure_distance}.}
                    \State Let $(\hat{i}, \hat{j}) = \underset{i, j \in 1, \dots, |S|, \ i \ne j}{argmin} D(i, j)$.
                    \State Call \texttt{Merge}($\hat{i}$, $\hat{j}$).

                \EndIf
    
            \EndProcedure

            \State
            
            \Procedure{Merge}{$i$, $j$}
                \State Let $s_k$ be a new structure with:
                \State \hspace{1em} $\mu_k$ from $\mu_i$ and $\mu_j$ using Equation~\ref{eq:merge:mu}
                \State \hspace{1em} $\Sigma_k$ from $\Sigma_i$ and $\Sigma_j$ using Equation~\ref{eq:cu}
                \State \hspace{1em} $w_k$ from $w_i$ and $w_j$ using Equation~\ref{eq:incremental:w}
                \State \hspace{1em} $T_k = T_i + T_j$
                \State Replace $s_i$ and $s_j$ in $S$ with $s_k$
            \EndProcedure

            \State

            \Procedure{GetClustering}{$\varepsilon$, $min\_pts$}
                \State Run DBSCAN($\varepsilon$, $min\_pts$) on $S$ using the distance
                \State \hspace{1em} function in Equation~\ref{eq:structure_distance}.
                \State \Return Cluster labels from DBSCAN.
            \EndProcedure
    
        \end{algorithmic}
    
    \end{algorithm}

\subsection{Algorithm Description}
\label{ssec:methods:description}

There is no initialization for SPC, the algorithm starts directly processing points from the stream.
There is, however, a burn-in phase during the first $n$ points, where $n$ is the maximum number of structures parameter, during which no structure
    merging occurs.
Each incoming point $x_i$, $i \le n$, during this phase is modeled by its own structure with mean $\mu_i = x$ and identity covariance (\textit{i.e.},
    Euclidean distance used in typicality computation).
One can still cluster the structures using DBSCAN as specified in Algorithm~\ref{alg:spc}, but examining the structures directly would not yet yield
    any useful information about the stream.

Even after $n$ structures have been created, the first step of processing a new point $x_j$, $j > n$, from the stream is to create a new structure
    with mean $x_j$ and identity covariance.
At this point, there are $n+1$ structures, so one must be removed.
This can happen in one of two ways.
If the weight of this structure is too small, the structure will either be simply deleted or merged into the best fit existing structure,
    depending on whether there exists another structure in which the NLT of the to-be-removed cluster is high enough.
If there are too many structures and all structures have too much weight to be deleted, the two structures that are deemed most similar according to
    Equation~\ref{eq:structure_distance} are merged.
    \section{Experiments}
\label{sec:experiments}

In this section, we evaluate SPC both qualitatively on synthetic datasets and quantitatively on real datasets against other state-of-the-art streaming
    clustering algorithms (CluStream~\cite{aggarwal2003clustream}, DenStream~\cite{cao2006density}, D-Stream~\cite{chen2007dstream},
    DBSTREAM~\cite{hahsler2016clustering}, and StreamSoNG~\cite{wu2021streamsong}.).
All parameters used in all experiments in this manuscript are documented in Table~\ref{tab:parameters}.
For plotting decision regions, we use the nearest neighbors algorithm and the distance function
\begin{equation}
    d(s_i, x_j) = 1 - \left( \frac{1}{1 + \left[ (x_j - \mu_i)^T \Sigma_i^{-1} (x_j - \mu_i) \right]^{\frac{1}{m-1}}} \right)^2
\end{equation}
where $s_i$ is the $i$-th structure tracked by SPC with mean $\mu_i$ and covariance $\Sigma_i$, and $x_j$ is an arbitrary point.

\begin{table}[]

    \begin{tabular}{llllllll}
        \hline
        \multicolumn{8}{c}{\textbf{Figure~\ref{fig:dataset:aggregation} - Aggregation Dataset}}                               \\ \hline
        \textbf{Parameter} & $n$ & $\gamma$ & $\beta$ & $m$ & $\varepsilon$ & $w_{min}$ & $NLT_{max}$   \\
        \textbf{Value}     & 30  & 0.0      & 0.0     & 1.5 & 0.95          & 0.01      & 3
    \end{tabular}

    \vspace{1em}

    \begin{tabular}{llllllll}
        \hline
        \multicolumn{8}{c}{\textbf{Figure~\ref{fig:dataset:sine} - Sine Dataset}}                                      \\ \hline
        \textbf{Parameter} & $n$ & $\gamma$ & $\beta$ & $m$ & $\varepsilon$ & $w_{min}$ & $NLT_{max}$   \\
        \textbf{Value}     & 30  & 0.1      & 0.05    & 1.4 & 0.95          & 0.01      & 3          
    \end{tabular}
    
    \vspace{1em}

    \begin{tabular}{llllllll}
        \hline
        \multicolumn{8}{c}{\textbf{Figure~\ref{fig:dataset:high_dim} - High Dimensionality Gaussians}}                                      \\ \hline
        \textbf{Parameter} & $n$ & $\gamma$ & $\beta$ & $m$ & $\varepsilon$ & $w_{min}$ & $NLT_{max}$   \\
        \textbf{Value}     & 50  & 0.0      & 0.00    & 1.5 & 0.95          & 0.01      & 3          
    \end{tabular}

    \begin{tabular}{llllllll}
        \hline
        \multicolumn{8}{c}{\textbf{Figure~\ref{fig:dataset:overlapping} - Overlapping Dataset}}                               \\ \hline
        \textbf{Parameter} & $n$ & $\gamma$ & $\beta$ & $m$ & $\varepsilon$ & $w_{min}$ & $NLT_{max}$   \\
        \textbf{Value}     & 30  & 0.0      & 0.0     & 1.5 & 0.95          & 0.01      & 3
    \end{tabular}

    \caption{Parameter values used for SPC experiments in this manuscript.}
    \label{tab:parameters}

\end{table}

\subsection{Synthetic Dataset}

\begin{figure*}[t!]
     
    \centering

    \begin{subfigure}{0.47\textwidth}
        \centering
        \includegraphics[width=0.9\textwidth]{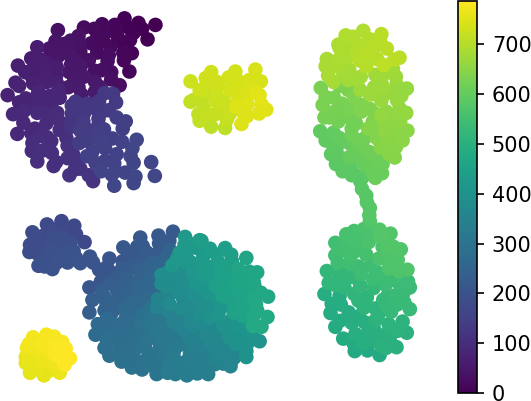}
        \caption{
            Dataset of seven clusters in two dimensions, colored by arrival time from blue (upper left cluster) to yellow (small lower left cluster).
        }
    \end{subfigure}
    \hspace{1em}
    \begin{subfigure}{0.47\textwidth}
        \centering
        \includegraphics[width=0.9\textwidth]{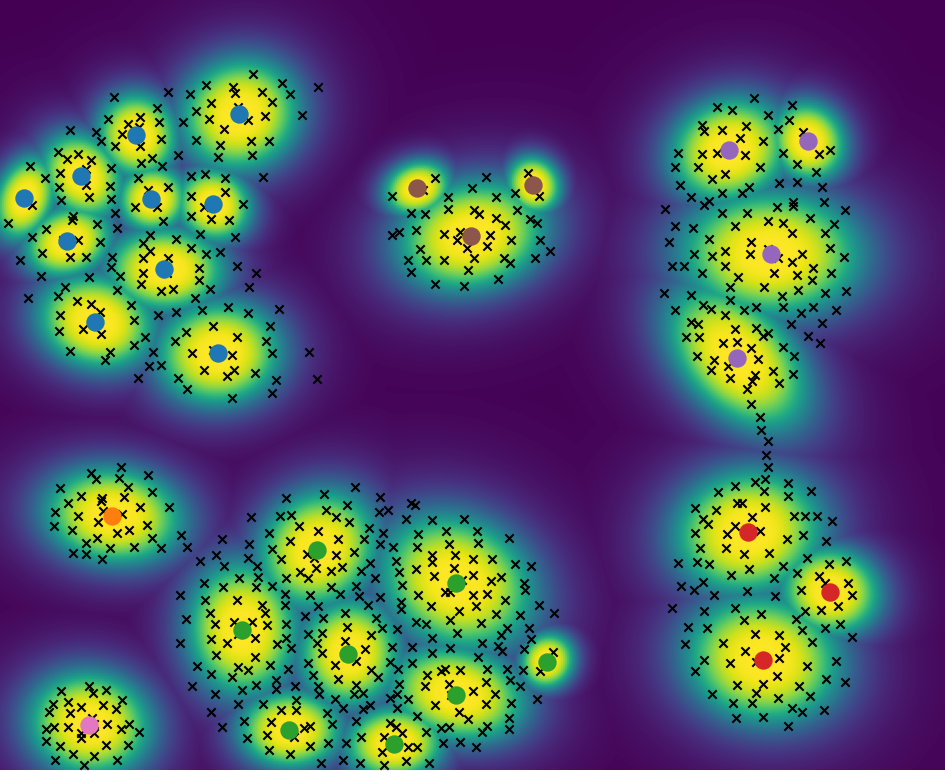}
        \caption{
            Final configuration of SPC with 30 structures.
            The final mean $\mu^{(T)}$ of each structure is plotted as a filled in circle along with the typicality induced by its covariance matrix.
            The color of each circle represents the cluster label assigned with DBSCAN, which we observe to be consistent with the true labels.
        }
        \label{fig:dataset:aggregation:clustering}
    \end{subfigure}

    \vspace{1em}

    \begin{subfigure}{0.47\textwidth}
        \centering
        \includegraphics[width=0.9\textwidth]{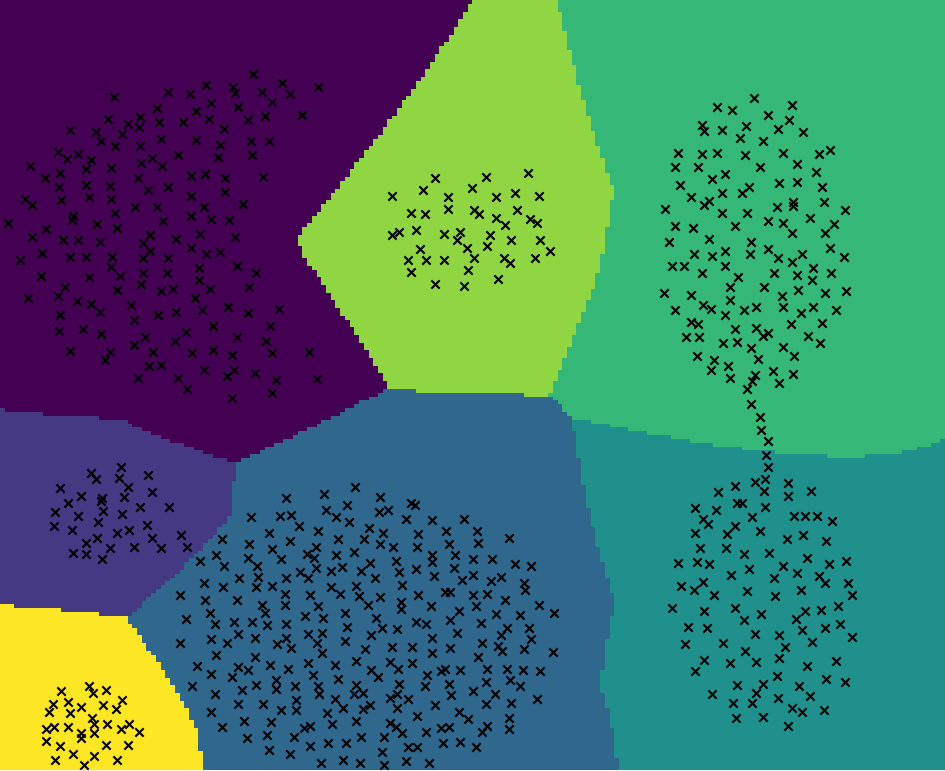}
        \caption{
            Decision region induced by running the nearest neighbor algorithm on a fine grid over the plot area.
        }
        \label{fig:dataset:aggregation:decision}
    \end{subfigure}
    \hspace{1em}
    \begin{subfigure}{0.47\textwidth}
        \centering
        \includegraphics[width=\textwidth]{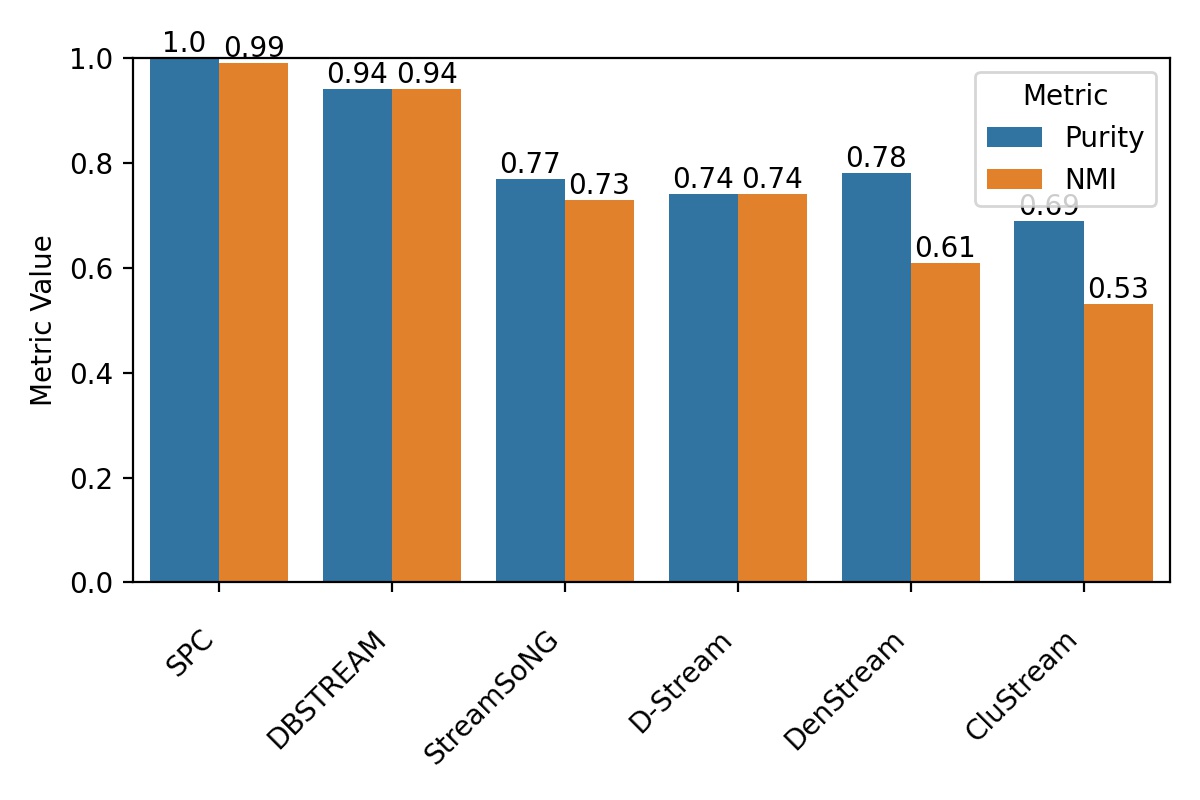}
        \caption{
            Comparison of streaming clustering algorithms on the synthetic dataset of Gionis et al.~\cite{gionis2007clustering} with respect to cluster
            purity and normalized mutual information (NMI).
        }
        \label{fig:dataset:aggregation:metrics}
    \end{subfigure}
   
    \caption{
        SPC performance on the synthetic clustering dataset of Gionis et al.~\cite{gionis2007clustering}.
    }
    
   \label{fig:dataset:aggregation}

\end{figure*}

We first evaluate SPC on a two-dimensional synthetic dataset with seven clusters, originally published by Gionis et al.~\cite{gionis2007clustering}.
Key challenges with this dataset include the non-Gaussian cluster in the upper left and the two sets of two clusters that bleed into each other.
SPC handled both challenges well and produced a very good clustering.
The decision region, shown in Figure~\ref{fig:dataset:aggregation:decision}, leaves nothing to be desired and arguably looks as if it was drawn by a
    human.

For this dataset, the forgetting factors $\gamma$ and $\beta$ were both set to zero so the algorithm weights the whole dataset equally, rather than
    placing more weight on recently observed points.
From the final position of the structures tracked by SPC (Figure~\ref{fig:dataset:aggregation:clustering}), it is clear that the entire dataset was
    effectively summarized.
Despite being computed over a single pass of the dataset, the final clustering is indistinguishable from iterative methods that are allowed to make
    as many passes over the dataset as required.

As expected from its high qualitative performance, SPC achieve a near-optimal value in both purity and NMI on this dataset, shown in
    Figure~\ref{fig:dataset:aggregation:metrics}.
DBSTREAM also performed very well on this dataset, followed by StreamSoNG, D-Stream, and DenStream.

\subsection{Nonstationary Dataset}

\begin{figure*}[t!]
     
    \centering

    \begin{subfigure}{0.47\textwidth}
        \centering
        \includegraphics[width=0.9\textwidth]{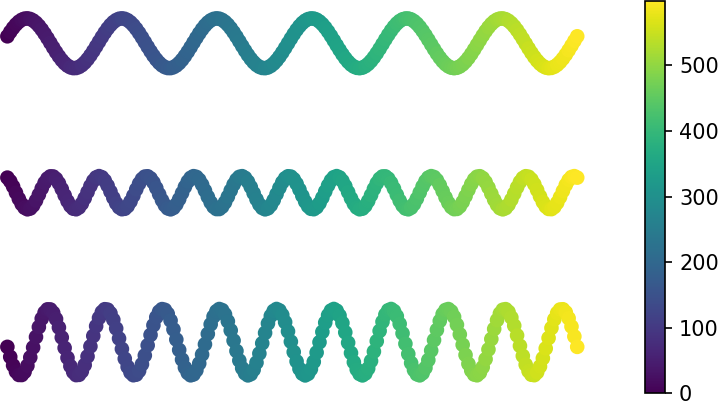}
        \caption{
            Dataset of three sine waves arriving simultaneously and moving unilaterally toward $+\infty$ on the $x$-axis, colored by arrival time
                from blue to yellow.
            Points in the stream are sampled in a round-robin pattern from each class.
        }
    \end{subfigure}
    \hspace{1em}
    \begin{subfigure}{0.47\textwidth}
        \centering
        \includegraphics[width=\textwidth]{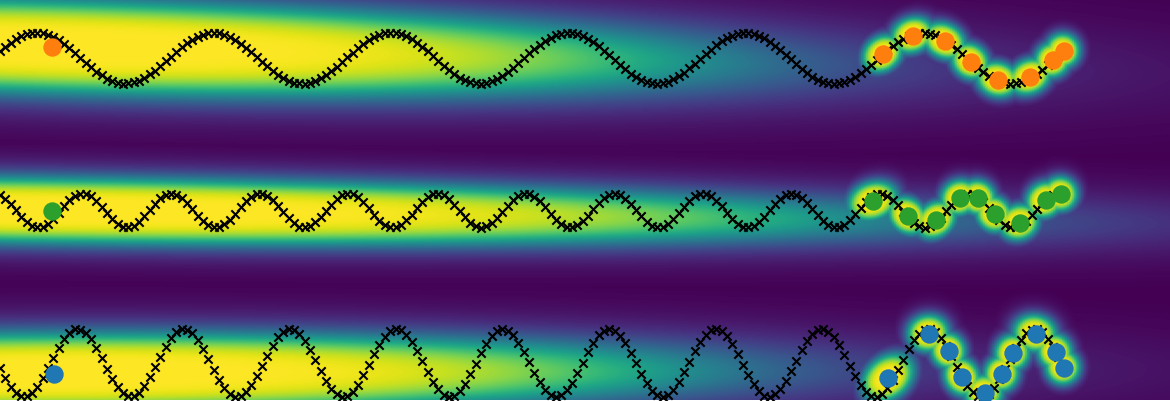}
        \caption{
            Final configuration of SPC on the sine wave dataset with 30 structures.
            The final mean $\mu^{(T)}$ of each structure is plotted as a filled in circle along with the typicality induced by its covariance matrix.
            The color of each circle represents the cluster label assigned with DBSCAN, which we observe to be consistent with the true labels.
        }
    \end{subfigure}

    \vspace{1em}

    \begin{subfigure}{0.47\textwidth}
        \centering
        \includegraphics[width=0.9\textwidth]{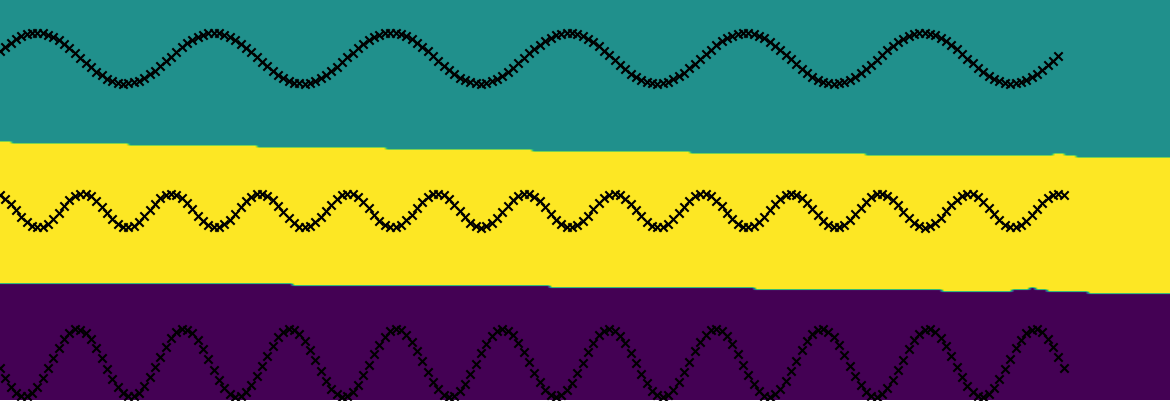}
        \caption{
            Decision region induced by running the nearest neighbors algorithm on a fine grid over the plot area.
        }
    \end{subfigure}
    \hspace{1em}
    \begin{subfigure}{0.47\textwidth}
        \centering
        \includegraphics[width=\textwidth]{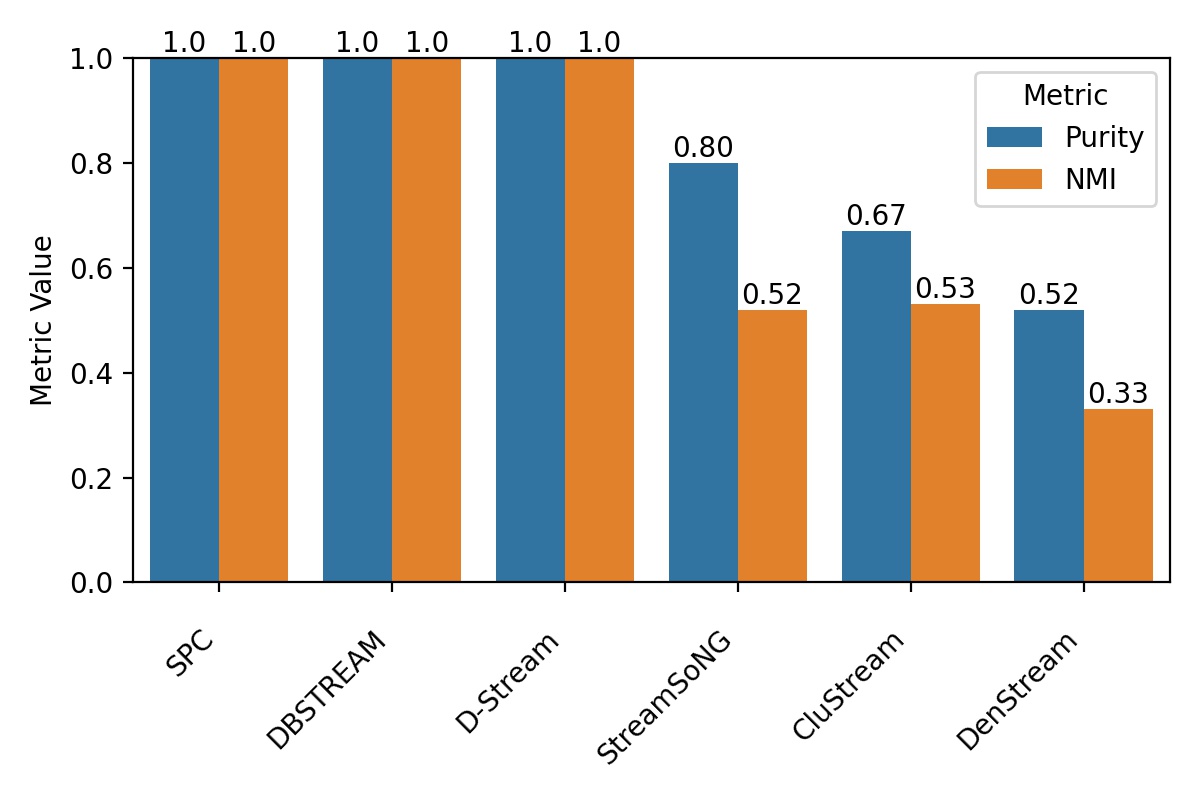}
        \caption{
            Comparison of streaming clustering algorithms on nonstationary sine dataset with respect to cluster purity and normalized mutual information
            (NMI).
        }
        \label{fig:dataset:sine:metrics}
    \end{subfigure}

    \caption{
        Nonstationary synthetic dataset of three sine waves illustrating the utility of SPC's forgetting factor in modeling newer data with finer detail.
        Old points are still modeled, but with less granularity.
        Here, SPC was run with a high forgetting factor of $\gamma=0.1$.
    }
    
   \label{fig:dataset:sine}

\end{figure*}

Datasets that have nonstationary clusters that continually evolve over time inevitably require clustering algorithms to either forget older points,
    create an unbounded number of clusters, or increase the number of points modeled by a single cluster.
SPC chooses the latter, as the first option is undesirable and the second violates the ``finite storage to model infinite data'' philosophy of
    streaming clustering.

In Figure~\ref{fig:dataset:sine}, we demonstrate a case where running SPC with a high forgetting factor is valuable.
This dataset has three highly non-Gaussian clusters, each modeled by a sine wave with different frequency and amplitude.
SPC was given a maximum of 30 structures to use in modeling this dataset, and allocated most of them to finely detail the most recently encountered
    points, appealing to the principle of spatiotemporal locality (new points in a stream are lie in close proximity to recently seen points).
One structure with a very large region of influence is used to model the early points in the stream from each sine wave.

We compare SPC against several other streaming clustering algorithms in Figure~\ref{fig:dataset:aggregation:metrics}.
Clearly, the nonstationary, non-Gaussian nature of this dataset was problematic for some algorithms.
SPC, D-Stream, and DBSTREAM achieved a perfect clustering, whereas CluStream, DenStream, and StreamSoNG struggled to either separate the three
    clusters or remember the whole stream.

When using DBSCAN to assemble structures into clusters, we use a specialized distance metric that leverages the covariance matrix of each structure.
We see that the very large structures modeling older points in the stream are given the same cluster label as the small structures modeling newer
    points in the stream, but that no two structures from different sine waves are given the same cluster label.
This is reflected in the perfect purity and NMI scores of Figure~\ref{fig:dataset:sine:metrics}.

\subsection{High Dimensionality Dataset}

\begin{figure}[t!]
     
    \centering
    \includegraphics[width=0.45\textwidth]{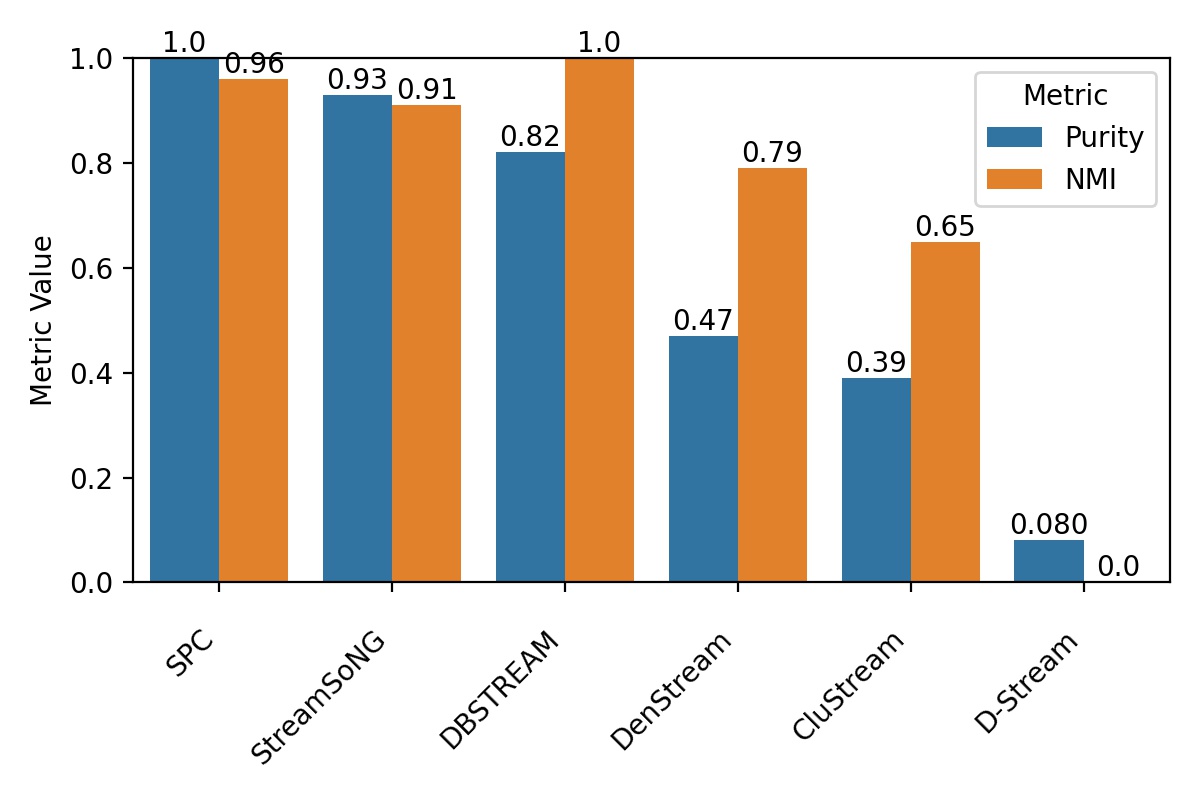}
    \caption{
        Comparison of streaming clustering algorithms on a high dimensional dataset~\cite{franti2006fast} of 1024 points from 16 Gaussian
            clusters in $\mathbb{R}^{1024}$ with respect to cluster purity and normalized mutual information (NMI).
    }
    
   \label{fig:dataset:high_dim}

\end{figure}

Lastly, we evaluate SPC on a very high dimensional dataset consisting of 1024 points from 16 well-separated Gaussians in 1024-dimensional
    space~\cite{franti2006fast}.
The results of running SPC and other comparison algorithms on this dataset are shown in Figure~\ref{fig:dataset:high_dim}.
We see that SPC, DBSTREAM, and StreamSoNG produce good sets of clusters on this dataset, with DenStream and CluStream attaining slightly lower
    performance.
We were not able to find a parameter set for D-Stream that produced a non-trivial clustering result.
It is well known that D-Stream does not scale to high dimensionality due to its need to produce a grid over the entire feature space, which, for
    1024-dimensional data, is highly intractable.

Notable drawbacks of covariance-based methods, like SPC, on high dimensional data are the difficulty in capturing the correlations between all pairs
    of variables and the quadratic (in dimensionality) storage requirement to retain full covariance matrices.
When covariance is not constrained in some way (\textit{i.e.}, spherical, diagonal, tied) to favor sparsity, there are $O(d^2)$ pairs of correlations
    in $d$-dimensional space, which can easily and overwhelmingly exceed the number of data points when $d$ is large.
The 1024-dimensional Gaussian data, for example, requires SPC to estimate over 1 million entries in each structure's covariance matrix, despite
    the stream only containing 1024 points.

In general, a stream in $d$ dimensions would need to have $O(n(1+d))$ elements for the memory consumption of SPC to be less than that of storing
the entire stream.
With $n=50$ structures in 1024-dimensional space, this would correspond to the stream containing on the order of 50,000 points.
In cases like this dataset, where it it doesn't hold that $n \gg d$, it is hard to motivate using SPC, or essentially any streaming clustering
algorithm, over static clustering algorithms that simply retain the stream and iterate over it repeatedly.

The purpose of this experiment is to show that, despite SPC's inherent incompatibility with high dimensional data, it can still
    perform well when clusters are sufficiently compact and well-separated.
As future work, we wish to extend SPC to maintain sparse, constrained estimates of the covariance in each structure so as to make it more practical
    in high dimensional use cases.
Generally, we would recommend employing some form of dimensionality reduction when applying SPC to such high dimensional streams.

\begin{figure*}[t!]
     
    \centering

    \begin{subfigure}{0.47\textwidth}
        \centering
        \includegraphics[width=0.9\textwidth]{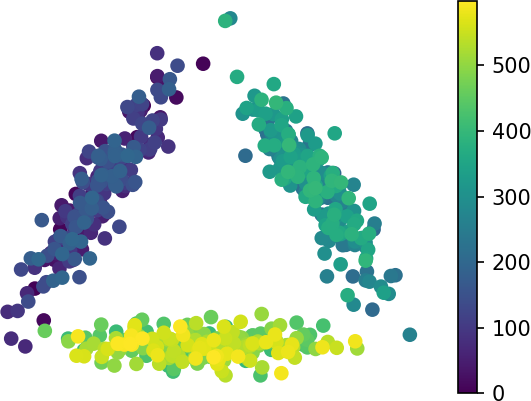}
        \caption{
            Dataset of three clusters in two dimensions, colored by arrival time.
            Clusters arrive sequentially, but within each cluster, points arrive in a random order.
        }
    \end{subfigure}
    \hspace{1em}
    \begin{subfigure}{0.47\textwidth}
        \centering
        \includegraphics[width=0.9\textwidth]{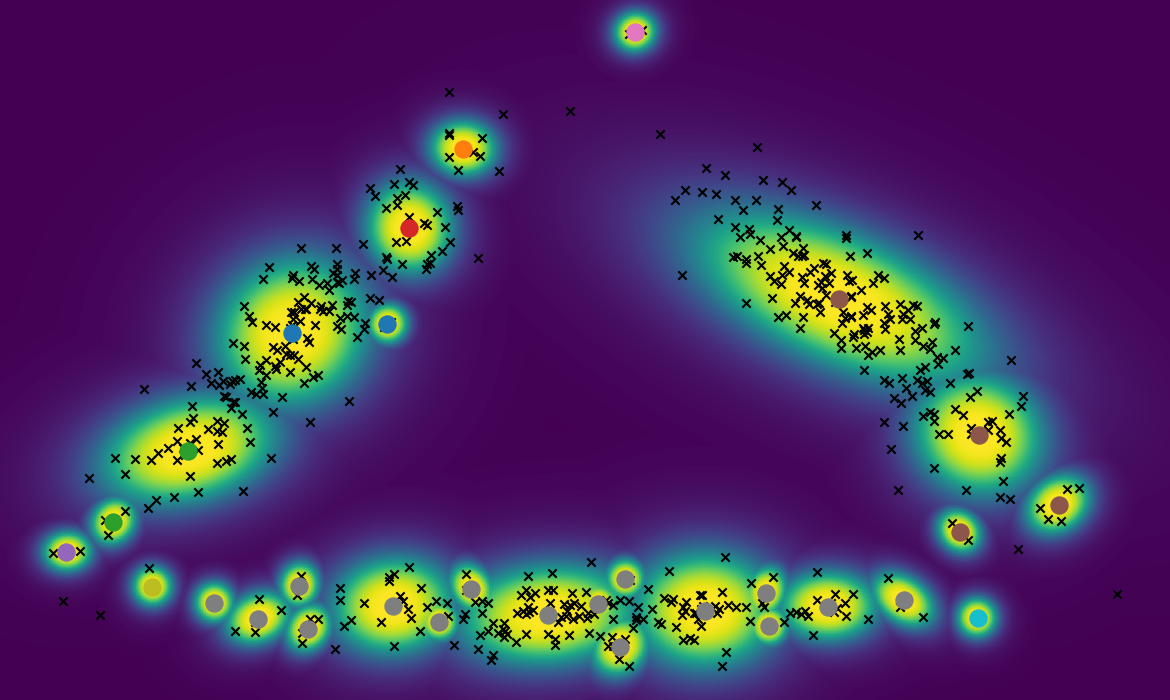}
        \caption{
            Final configuration of SPC with 30 structures.
            The final mean $\mu^{(T)}$ of each structure is plotted as a filled in circle along with the typicality induced by its covariance matrix.
            The color of each circle represents the cluster label assigned with DBSCAN, which we observe to be mostly consistent with the true labels.
        }
        \label{fig:dataset:overlapping:clustering}
    \end{subfigure}

    \vspace{1em}

    \begin{subfigure}{0.47\textwidth}
        \centering
        \includegraphics[width=0.9\textwidth]{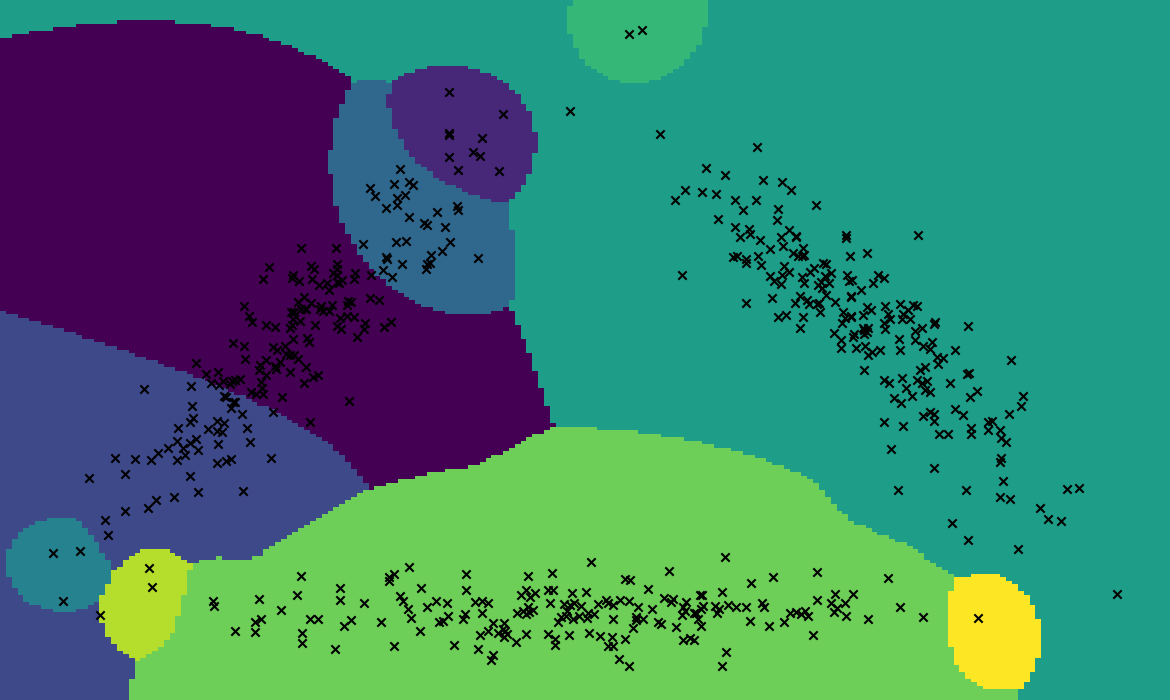}
        \caption{
            Decision region induced by running the nearest neighbor algorithm on a fine grid over the plot area.
        }
        \label{fig:dataset:overlapping:decision}
    \end{subfigure}
    \hspace{1em}
    \begin{subfigure}{0.47\textwidth}
        \centering
        \includegraphics[width=\textwidth]{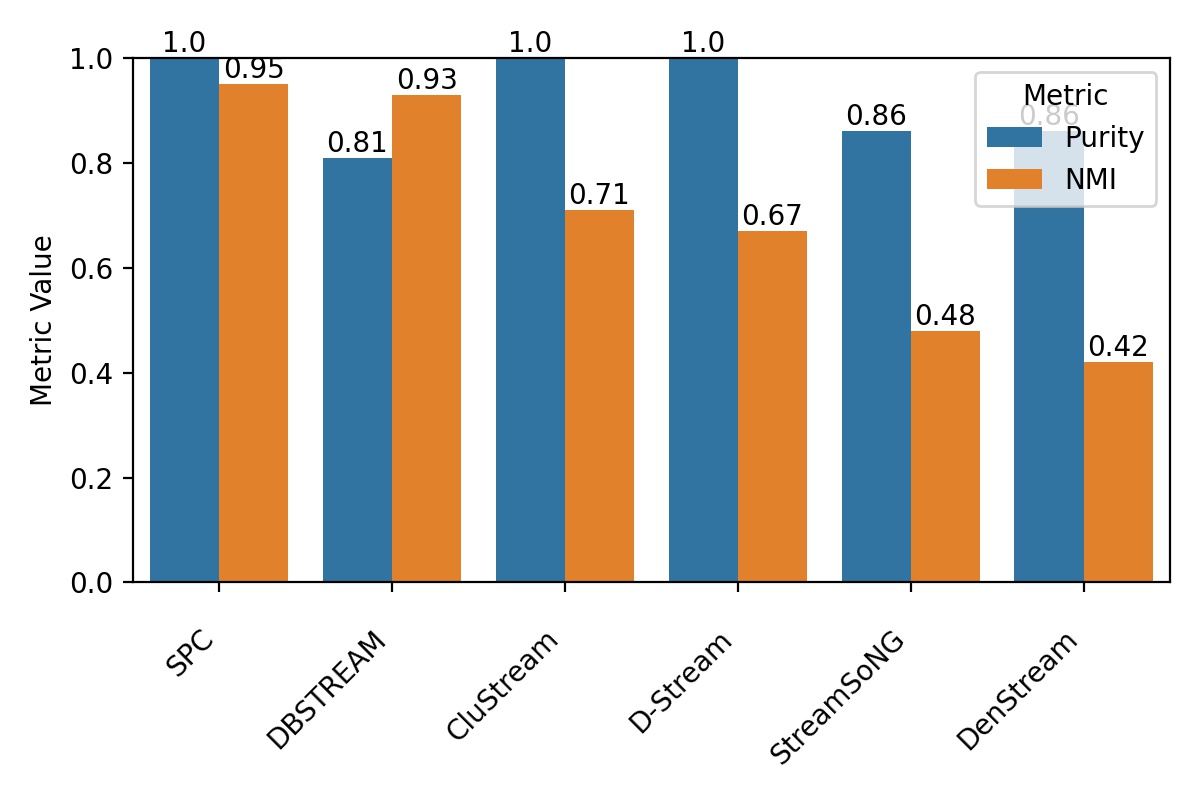}
        \caption{
            Comparison of streaming clustering algorithms on an overlapping cluster dataset with respect to cluster
            purity and normalized mutual information (NMI).
        }
        \label{fig:dataset:overlapping:metrics}
    \end{subfigure}
   
    \caption{
        SPC performance on an overlapping cluster dataset.
    }
    
   \label{fig:dataset:overlapping}

\end{figure*}

\subsection{Overlapping Dataset}

Datasets with overlapping clusters present a challenge for most streaming clustering algorithms, SPC included.
To evaluate SPC and related algorithms on overlapping data, we created a synthetic dataset of three highly correlated Gaussians in the shape
    of a triangle.
This dataset, shown in Figure~\ref{fig:dataset:overlapping}, tests the ability of algorithms to separate clusters that are connected by
    low-density regions of points.
The most common pitfall when running streaming clustering algorithms on this dataset is grouping the entire dataset into a single cluster,
    making many algorithms sensitive to parameter choice.

As observed in Table~\ref{tab:parameters}, we use the same SPC parameters for this dataset as we do for the dataset in
    Figure~\ref{fig:dataset:aggregation}, though we did have to adjust the parameters for the other algorithms.
Clearly, this dataset was challenging across the board, evidenced by universally lower metrics in Figure~\ref{fig:dataset:overlapping:metrics}.
SPC still had the highest performance\footnote{
    SPC achieved a perfect purity of 1.0 in Figure~\ref{fig:dataset:overlapping:metrics} despite clearly having a cluster partition that is not
    identical to the ground truth partition because the clusters in SPC are all homogeneous with respect to ground truth labels, which is why we also
    evaluate based on NMI.
}, but in looking at the decision region shown in Figure~\ref{fig:dataset:overlapping:decision}, we see that
    the leftmost cluster was split into three sizeable clusters and that there are several remaining structures placed on outlier points.

This dataset is also valuable in illustrating the complex decision regions that can be induced by the typicality function in
    Equation~\ref{eq:mahal_typicality} (Figure~\ref{fig:dataset:overlapping:decision}).
As we get far away all of the points in the dataset, we see that the nearest neighbor structure is actually the structure with largest covariance
    rather than the structure whose mean is closest in terms of Euclidean distance, although typicality is extremely low in all structures.
    \section{Conclusions}
\label{sec:conclusion}

We have proposed a single-pass possibilistic clustering algorithm, coined SPC, that demonstrates strong performance when applied to a wide array of
    clustering datasets.
Key contributions of SPC include the use Mahalanobis distance to compute typicality, merging of structures over a damped window, and the transfer of
    covariance union from the domain of multiple hypothesis tracking to that of streaming clustering.

A key design choice in SPC is its ease of application to new datasets; the only data-dependent parameter is the fuzzifier $m$, which, in most cases,
    can remain around $m=1.5$.
The decay factors $\gamma$ and $\beta$ allow the user to balance long term memory of the stream with prioritization of recently seen points, enabling
    SPC to function effectively on both stationary streams ($\gamma = \beta = 0$) and nonstationary streams ($\gamma, \beta > 0$), all while
    maintaining a constant memory footprint.

SPC either exceeds the performance of, or is competitive with, five other streaming clustering algorithms from the literature.
Qualitatively speaking, SPC produces extremely high quality decision regions on two dimensional data, and, quantitatively speaking, achieves high
    purity and NMI on a compact and well-separated dataset in 1024 dimensions.

    \printbibliography[title={References}]
    \balance

\end{document}